\documentclass[10pt,twocolumn,letterpaper]{article}

\usepackage{iccv}
\usepackage{times}
\usepackage{epsfig}
\usepackage{graphicx}
\usepackage{amsmath}
\usepackage{amssymb}

\usepackage{colortbl}

\definecolor{gbypink}{rgb}{0.99, 0.91, 0.95}
\definecolor{gbygray}{rgb}{0.5,0.9,0.9}
\usepackage{threeparttable}
\usepackage{color}
\usepackage[colorlinks,linkcolor=blue]{hyperref}

\iccvfinalcopy 

\def\iccvPaperID{4195} 

\ificcvfinal\fi

\begin{document}

\title{Self-supervised Monocular Depth Estimation for All Day Images using \\ Domain Separation}

\author{Lina Liu$^{1,2}$, 
    Xibin Song$^{2,4}$\thanks{Corresponding authors}, 
    Mengmeng Wang$^{1}$, 
    Yong Liu$^{1,3*}$ and 
    Liangjun Zhang$^{2,4}$\\
    
    \textsuperscript{\rm 1}Institute of Cyber-Systems and Control, Zhejiang University, China \\ 
    \textsuperscript{\rm 2}Baidu Research, China   \quad
    \textsuperscript{\rm 3}Huzhou Institue of Zhejiang University, China \\
    \textsuperscript{\rm 4}National Engineering Laboratory of Deep Learning Technology and Application, China
    \\
{\tt\footnotesize \{linaliu,mengmengwang\}@zju.edu.cn,song.sducg@gmail.com,liangjunzhang@baidu.com,yongliu@iipc.zju.edu.cn}
}


\maketitle
\ificcvfinal\thispagestyle{empty}\fi

\begin{abstract}

Remarkable results have been achieved by DCNN based self-supervised depth estimation approaches. However, most of these approaches can only handle either day-time or night-time images, while their performance degrades for all-day images due to large domain shift and the variation of illumination between day and night images. To relieve these limitations, we propose a domain-separated network for self-supervised depth estimation of all-day images. Specifically, to relieve the negative influence of disturbing terms (illumination, etc.), we partition the information of day and night image pairs into two complementary sub-spaces: private and invariant domains, where the former contains the unique information (illumination, etc.) of day and night images and the latter contains essential shared information (texture, etc.). Meanwhile, to guarantee that the day and night images contain the same information, the domain-separated network takes the day-time images and corresponding night-time images (generated by GAN) as input, and the private and invariant feature extractors are learned by orthogonality and similarity loss, where the domain gap can be alleviated, thus better depth maps can be expected. Meanwhile, the reconstruction and photometric losses are utilized to estimate complementary information and depth maps effectively. Experimental results demonstrate that our approach achieves state-of-the-art depth estimation results for all-day images on the challenging Oxford RobotCar dataset, proving the superiority of our proposed approach. Code and data split are available at \href{https://github.com/LINA-lln/ADDS-DepthNet}{https://github.com/LINA-lln/ADDS-DepthNet}.

\end{abstract}



\vspace{-0.2in}
\section{Introduction}\label{sec:introduction}

Self-supervised depth estimation has been applied in a wide range of fields such as augmented reality~\cite{azuma1997survey}\cite{carmigniani2011augmented}, 3D reconstruction~\cite{izadi2011kinectfusion}, SLAM~\cite{liu2021fcfr}\cite{mur2017orb}\cite{pritsker1984introduction} and scene understanding~\cite{armeni2017joint}\cite{chen2019towards} since it does not need large and accurate ground-truth depth labels as supervision. The depth information can be estimated by the implicit supervision provided by the spatial and temporal consistency present in image sequences. Benefiting from the well developed deep learning technology, impressive results have been achieved by Deep Convolution Neural Network(DCNN) based approaches~\cite{Guizilini_2020_CVPR}\cite{liu2021learning}\cite{long2020multi}\cite{long2020occlusion}, which outperform traditional methods that rely on handcrafted features and exploit camera geometry and/or camera motion for depth and pose estimation.



\begin{figure}[t!]
	\centering
	\includegraphics[width=1\linewidth]{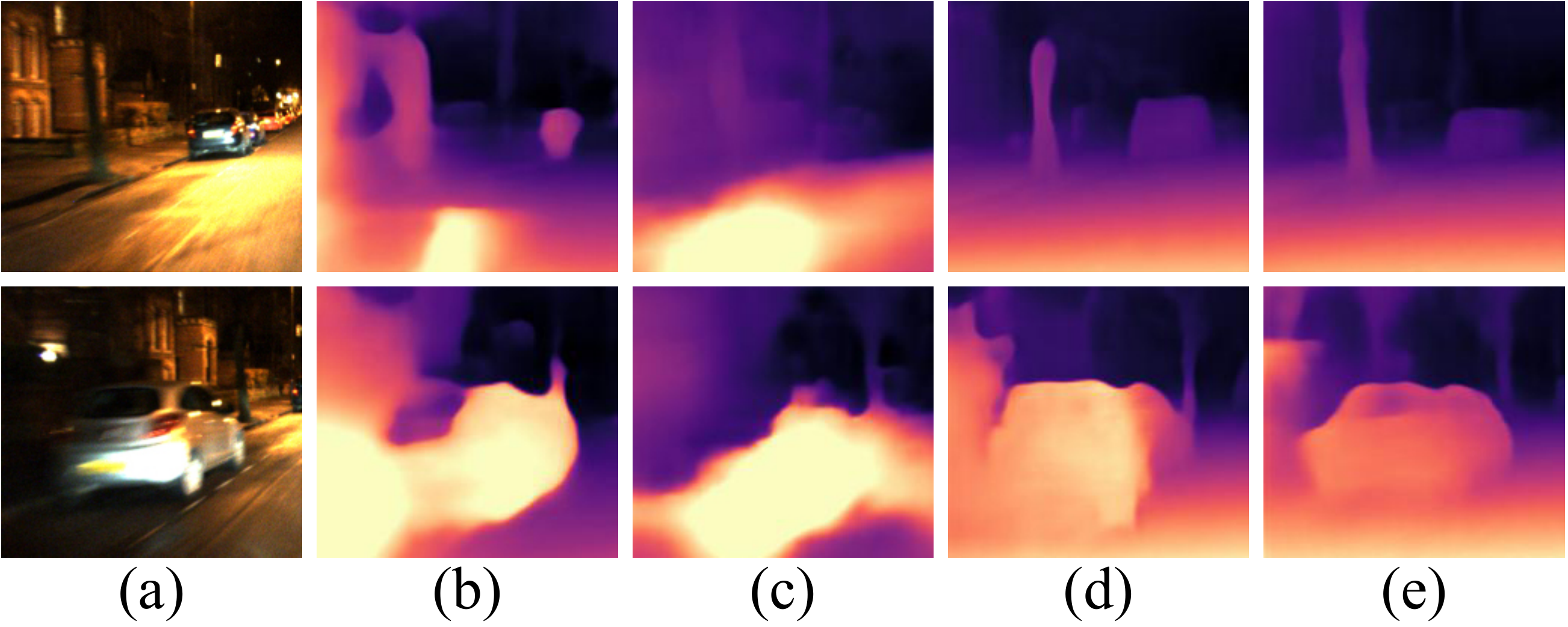}
	
	\caption{Comparison with other approaches on Oxford RobotCar dataset~\cite{maddern20171}. From left to right: (a) Night Images, (b) Monodepth2~\cite{godard2019digging}, (c) HR-Depth~\cite{lyu2020hr}, (d) Monodepth2+CycleGAN~\cite{CycleGAN2017}, (e) Ours. }
	\label{fig:fig_teaset}
	\vspace{-0.2in}
\end{figure}



However, most of current DCNN based self-supervised depth estimation approaches~\cite{Johnston_2020_CVPR}\cite{Wang_2020_CVPR}\cite{Watson_2019_ICCV}\cite{Zhou_2019_ICCV} mainly solve the problem of depth estimation on day-time images, which are evaluated by day-time benchmarks, such as KITTI~\cite{geiger2013vision} and Cityscapes~\cite{cordts2016cityscapes}. They fail to generalize well on all-day images due to the large domain shift between day and night images. The night-time images are unstable due to the low visibility and non-uniform illumination arising from multiple and moving lights. Methods~\cite{im2018robust}\cite{kim2018multispectral} are proposed by applying a commonly used depth estimation strategy for images captured in low-light conditions. However, the performance is limited due to the unstable visibility. Meanwhile, generative adversarial networks(GAN), such as CycleGAN~\cite{CycleGAN2017}, are also used to solve the problem of depth estimation on night-time images by translating information of night-time to day-time in both image levels and feature levels. Unfortunately, due to the inherent domain shift between day and night-time images, it is difficult to obtain natural day-time images or features with GAN using night-time images as input, thus the performance is also limited. Fig.~\ref{fig:fig_teaset} (b) and (c) demonstrate the results of Monodepth2~\cite{godard2019digging} and HR-Depth~\cite{lyu2020hr} of night-time images. Monodepth2\cite{godard2019digging} is an effective self-supervised depth estimation approach, and HR-Depth\cite{lyu2020hr} make a series of improvements based on Monodepth2\cite{godard2019digging}. Fig.~\ref{fig:fig_teaset} (d) demonstrate the result Monodepth2~\cite{CycleGAN2017} with CycleGAN translated image as input. We can see that the depth details are failed to be estimated due to the non-uniform illumination of night-time images.

For a scene in real-world, the depth is constant if the viewpoint is fixed, while the disturbing terms, such as illumination, varies as time goes, which will disturb the performance of self-supervised depth estimation, especially for night-time images.~\cite{Dijk_2019_ICCV} also proves that texture information plays more important roles on depth estimation than exact color information. To cater to the above issues, we propose a domain-separated network for self-supervised depth estimation of all-day RGB images. The information of day and night image pairs are separated into two complementary sub-spaces: private and invariant domains. Both domains use DCNN to extract features. The private domain contains the unique information (illumination, etc.) of day and night-time images, which will disturb the performance of depth estimation. In contrast, the invariant domain contains invariant information (texture, etc.), which can be used for common depth estimation. Thus the disturbed information can be removed and better depth maps will be obtained. 

Meanwhile, unpaired day and night images always contain inconsistent information, which interferes with the separation of private and invariant features. Therefore, the domain-separated network takes a paired of the day-time image and corresponding night-time image (generated by GAN) as input, the private and invariant feature extractors are first utilized to extract private (illumination, etc.) and invariant (texture, etc.) features using orthogonality and similarity losses, which can obtain more effective features for depth estimation of both day and night-time images. Besides, constraints in feature and gram matrices levels are leveraged in orthogonality losses to alleviate the domain gap, thus more effective features and fine-grain depth maps can be obtained. Then, depth maps and corresponding RGB images are reconstructed by decoder modules with reconstruction and photometric losses. Note that real-world day-time and night-time images can be tested directly. As shown in Fig.~\ref{fig:fig_teaset} (e), our approach can effectively relieve the problems of low-visibility and non-uniform illumination, and achieves more appealing results for night-time images. 

The main contributions can be summarized as:

\begin{itemize}
    \item We propose a domain-separated framework for self-supervised depth estimation of all-day images. It can relieve the influence of disturbing terms in depth estimation by separating the all-day information into two complementary sub-spaces: private (illumination, etc.) and invariant (texture, etc.) domains, thus better depth maps can be expected;
    
    \item Private and invariant feature extractors with orthogonality and similarity losses are utilized to extract effective and complementary features to estimate depth information. Meanwhile, the reconstruction loss is employed to refine the obtained complementary information (private and invariant information);
    
    \item Experimental results on the Oxford RobotCar dataset demonstrate that our framework achieves state-of-the-art depth estimation performance for all-day images, which confirms the superiority of our approach.
    
\end{itemize}

\section{Related work}


\subsection{Day-time Depth Estimation}




Self-supervised depth estimation has been extensively studied in recent years. \cite{zhou2017unsupervised}\cite{godard2017unsupervised} are the first self-supervised monocular depth estimation approaches which train the depth network along with a separate pose network. Meanwhile, \cite{godard2019digging}\cite{Gordon_2019_ICCV}\cite{guizilini20203d}\cite{klingner2020selfsupervised} make a series of improvements for outdoor scenes, which are sufficiently evaluated on KITTI dataset~\cite{geiger2013vision} and Citycapes dataset~\cite{cordts2016cityscapes} subsequently. \cite{long2021adaptive}\cite{Ramamonjisoa_2020_CVPR} \cite{IndoorSfMLearner} outperform better results in indoor scenes.

KITTI~\cite{geiger2013vision} and Citycapes~\cite{cordts2016cityscapes} datasets only contain day-time images, and all of the above methods are excellently improved for these scenes. However, the self-supervised depth estimation approaches for all-day images have not been well addressed before, and the performance of current approaches on night-time images is limited due to the low-visibility and non-uniform illuminations.


\subsection{Night-time Depth Estimation}



\begin{figure*}[htb]
	\centering
	\includegraphics[width=1\linewidth]{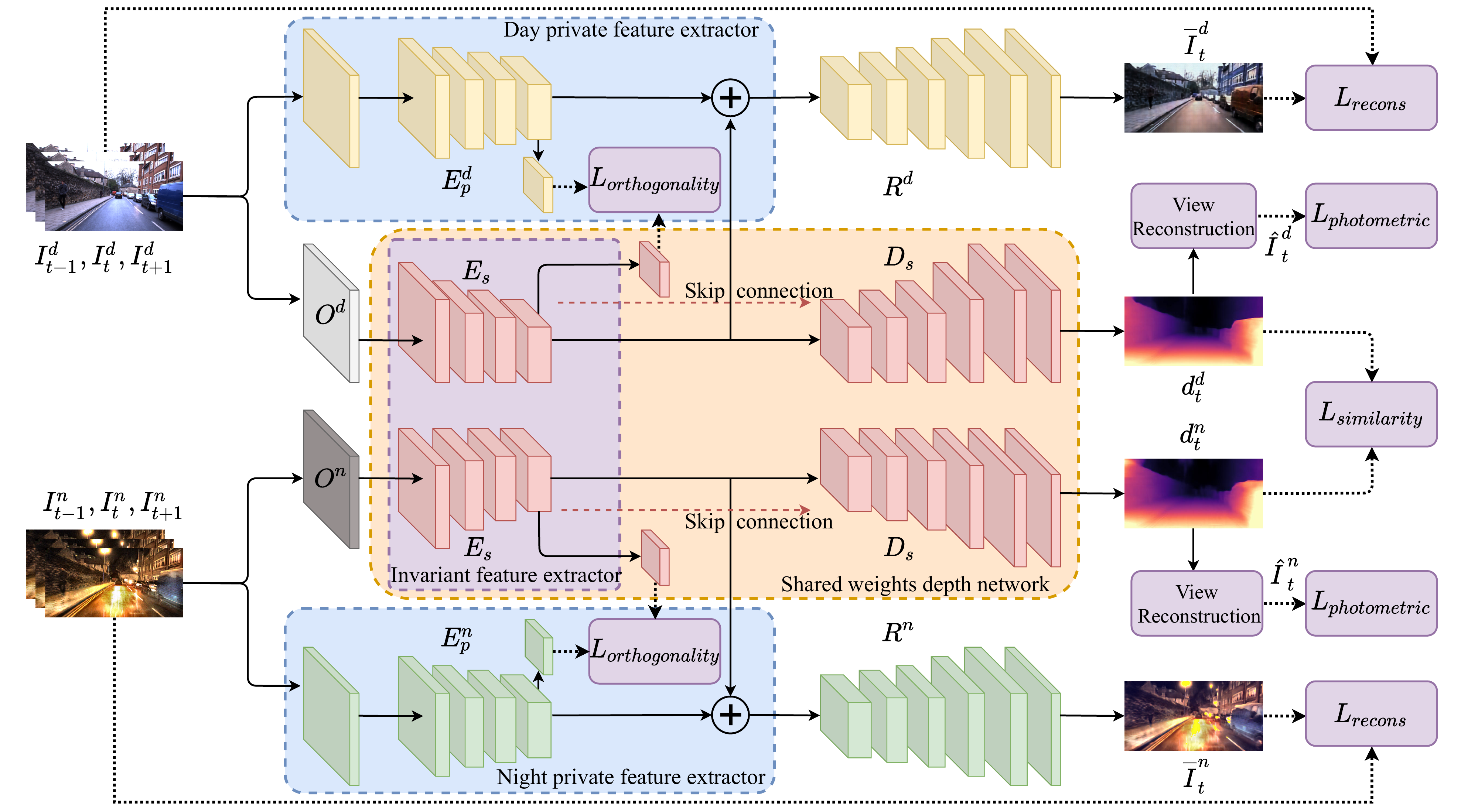}
	
	\caption{Overview of the network architecture. The network architecture includes three parts: Shared weights depth network (orange area), Day private branch (yellow structure) and night private branch (green structure). Day-time and night-time images are the input of the shared weights depth network, which extracts the invariant features first, and then estimates corresponding depth maps. Meanwhile, the day private feature extractor and night private feature extractor (blue area) extract the private features of day and night, respectively, which are constrained by orthogonality loss to get complementary features. And the private and invariant features are added to reconstruct the original input images with the reconstruction loss. In inference, only operations of $O^d$, $O^n$ and shared weights depth network are used to estimate depth.}
	\label{fig:fig_pipeline}
	\vspace{-0.2in}
\end{figure*}



Approaches have also been proposed for self-supervised depth estimation of night-time images.

\cite{kim2018multispectral}\cite{lu2021alternative} propose to use additional sensors to estimate depth of night-time images. To estimate all day time depth, \cite{kim2018multispectral} utilizes a thermal imaging camera sensor to reduce the influence of low-visibility in the night-time, while \cite{lu2021alternative} adds LiDAR to provide additional information in estimating depth maps at night-time. Meanwhile, using generate adversarial network, \cite{sharma2020nighttime} and \cite{vankadari2020unsupervised} propose effective strategies for depth estimation of night-time images. \cite{sharma2020nighttime} utilizes a translation network with light effects and uninformative regions that can render realistic night stereo images from day stereo images, and vice versa. During inference, a separate network during the day and night is used to estimate the stereo depth of night-time images. \cite{vankadari2020unsupervised} proposes an adversarial domain feature adaptation method to reduce the domain shift between day and night images at the feature level. Finally, independent encoders and a shared decoder are used for the day-time and night-time images during inference.

Though remarkable progress has been achieved, due to the large domain shift between day-time and night-time images, it is difficult to obtain natural day-time images or features with night-time images as input, thus the performances of these approaches are limited.



\subsection{Domain Adaptation}

Most depth estimation and stereo matching domain adaptation methods mainly focus on the migration between the synthetic domain and the real domain or between different datasets. Most methods usually translate images from one domain to another. To reduce the requirements of real-world images in depth estimation, \cite{atapour2018real}\cite{zhao2019geometry}\cite{zhao2020domain} explore image translation techniques to generate synthetic labeled data. \cite{maximov2020focus} tackles synthetic to real depth estimation issue by using domain invariant defocus blur as direct supervision. \cite{zhang2020domain} proposes a domain normalization approach of stereo matching that regularizes the distribution of learned representations to allow them to be invariant to domain differences.

Compared with previous approaches, we propose an effective domain separation framework for all-day self-supervised depth estimation, which can effectively handle the problem of domain shift between day and night-time images.

\section{Approach}

We propose a domain-separated framework to relieve the influence of disturbing terms, which takes day-time images and corresponding night-time images generated by GAN as input, and Fig.~\ref{fig:fig_pipeline} demonstrates the pipeline of our proposed domain separated framework. 



\subsection{Domain Separated Framework}


For day-time and night-time images of a scene, the depth information should be consistent, though the illumination of these image pairs is quite different. This means that the essential information of corresponding day-time images and night-time images of a scene should be similar. Here, we separate the information of day and night-time images into two parts:


\vspace{-0.2in}

\begin{equation}
    \begin{split}
        I^{d} &= I^{d}_{i}+I^d_{p}, \\
        I^{n} &= I^{n}_{i}+I^n_{p} 
    \end{split}
\end{equation}
where $I^d_{i}$ and $I^n_{i}$ mean the invariant information of day and night images, which should be similar of the same scene, and $I^d_{p}$ and $I^n_{p}$ mean the different private information (illumination, etc.) of day and night images, respectively. 

Inspired by~\cite{Dijk_2019_ICCV}, the illumination of a scene is different as time goes, while the depth of the scene is constant, thus the illumination components ($I^d_{p}$ and $I^n_{p}$) of a scene play fewer roles in self-supervised depth estimation. As shown in Fig.~\ref{fig:fig_pipeline}, the proposed domain-separated framework separates the images into two complementary sub-spaces in feature levels (elucidated in Fig.\ref{fig_feature_visual}), and the invariant components are utilized for depth estimation. 

Moreover, it is quite difficult to guarantee that the real-world day-time and night-time images of a scene contain the same information except for the private information (illumination, etc.), since there are always moving objects in outdoor scenes. This will mislead the network to obtain private and invariant components of images. Therefore, CycleGAN\cite{CycleGAN2017} is used to translate day-time images to night-time images, where the day-time and corresponding generated night-time images are regarded as input image pairs. It ensures that the invariant information is consistent, and all objects are in the same position, reducing the loss of essential information during the process of separating private information. Note that other GANs can also be used here.

Inspired by \cite{bousmalis2016domain}, our domain-separated framework uses two network branches to extract the private and invariant information of an image in feature levels, respectively. Given the input day image sequences $\{I^d_{t-1}, I^d_{t}, I^d_{t+1}\}$ and the corresponding generated night images sequences $\{I^n_{t-1}, I^n_{t}, I^n_{t+1}\}$, where $t$ represents the t-th frame image arranged in chronological order, the day private feature extractor $E^d_{p}$ and night private feature extractor $E^{n}_{p}$ are used to extract private features of day-time images and night-time images $f^{d}_{p}$ and $f^{n}_{p}$, respectively. The invariant feature extractor $E^{d}_{i}$ and $E^{n}_{i}$ are utilized to extract invariant features of day-time and night-time images $f^{d}_{i}$ and $f^{n}_{i}$, respectively. Since the input day-time and night-time images contains same essential information, $E^{d}_{i}$ and $E^{n}_{i}$ are weight-shared, which is defined as $E_{s}$. Then the feature extraction process can be formulated as:

\vspace{-0.2in}
\begin{equation}
    \begin{split}
        {f^{d}_{p_t}} &= E^{d}_{p}(I^d_{t}), {f^{n}_{p_t}} = E^{n}_{p}(I^n_{t}) \\
        {f^d_{i_t}} &= E_{s}(I^d_{t}), {f^n_{i_t}} = E_{s}(I^n_{t}) \\
    \end{split}
\end{equation}
where t in the subscript denotes the t-th frame of day and night-time. 


Then, decoders are used to reconstruct the corresponding depth maps of day and night-time images. As shown in Fig.~\ref{fig:fig_pipeline}, the red decoder $D_{s}$ represents the depth recovery module of shared weights depth network, and the yellow decoder $R^{d}$ and green decoder $R^{n}$ denote the reconstructed feature restoration branch. The process of the depth map and image reconstruction can be formulated as:

\vspace{-0.2in}
\begin{equation}
\vspace{-0.1in}
    \begin{split}
        \overline{I}^d_{t} &= R^{d}(f^d_{p_t}+f^d_{i_t}) \\
        \overline{I}^n_{t} &= R^{n}(f^n_{p_t}+f^n_{i_t}) \\
        d^d_{t} &= D_{s}(f^d_{i_t}) \\
        d^n_{t} &= D_{s}(f^d_{i_t})
    \end{split}
\end{equation}
where $\overline{I}^d_{t}$ and $\overline{I}^n_{t}$ are the reconstructed images of t-th frame by $R^{d}$ and $R^{n}$, and $d^d_{t}$ and $d^n_{t}$ are the corresponding depth maps estimated by $D_s$.

\vspace{-0.05in}
\subsection{Loss function}

To obtain private and invariant features and well estimate depth information of all-day images in a self-supervised manner, different losses are leveraged here, including reconstruction loss, similarity loss, orthogonality loss and photometric loss.
\vspace{-0.1in}
\subsubsection{Reconstruction Loss}

The private and invariant features are complementary information which can be used to reconstruct the original RGB images. Hence, we use reconstruction loss to refine the domain separated framework, which is defined as:

\vspace{-0.2in}
\begin{equation}
    \vspace{-0.1in}
    \begin{split}
        L_{recons} &= \frac{1}{N}\sum_x (\overline{I}^d_{tx}-I^d_{tx})^2+ \frac{1}{N^2}(\sum_x (\overline{I}^d_{tx}-I^d_{tx}))^2\\
        &+ \frac{1}{N}\sum_x (\overline{I}^n_{tx}-I^n_{tx})^2+ \frac{1}{N^2}(\sum_x (\overline{I}^n_{tx}-I^n_{tx}))^2
    \end{split}
\end{equation}
where $x \in [1,N]$, $N$ is the pixel number of $I^n_{t}$ and ${I}^d_{t}$.
\subsubsection{Similarity Loss}


The proposed domain separated framework takes day-time images and corresponding generated night-time images (CycleGAN~\cite{CycleGAN2017}) as input, and the estimated depth maps of day-time and night-time images should be consistent. Due to the inherent advantages of the day-time image in depth estimation, the estimated depth of the night image is expected to be as close as possible to the day-time, that is, the depth of the day-time image is used as a pseudo-label to constrain the depth of the night-time image. So the similarity loss is defined as:


\vspace{-0.2in}
\begin{equation}
    \begin{split}
        L_{simi} &= \frac{1}{N}\sum_x (d^n_{tx}-\widetilde{d}^d_{tx})^2 
    \end{split}
\end{equation}
where $x \in [1,N]$, $N$ is the pixel number of $d^n_{t}$ and ${d}^d_{t}$, $x$ is the x-th pixel. $\widetilde{d}^d_{t}$ means that the gradient of ${d}^d_{t}$ is cut off during back propagation. 



\subsubsection{Orthogonality Loss}

As discussed above, the private and invariant features of an image are complementary and completely different. Therefore, two types of orthogonality losses are utilized here to guarantee the private and invariant features are completely different.



\textbf{Direct feature orthogonality loss:} the private and invariant feature extractors obtains 3-D private and invariant features ($f_{p}$ and $f_{i}$) which have large sizes. To reduce the complexity, we first use a convolution layer with a kernel size of $1\times 1$ to reduce the size of obtained private and invariant features ($v_p$ and $v_i$), then we straighten the reduced features into 1-D feature vectors. Finally, we calculate the inner product (orthogonality loss) between the private and invariant feature vectors, which is defined as $L_{f}$.

\textbf{Gram matrices orthogonality loss:} inspired by style transfer \cite{ghiasi2017exploring}, Gram matrix is widely used in style transfer to represent the style of the features. The private and invariant features can be considered to have different styles. Hence, we first calculate the Gram matrices ($\eta_p$ and $\eta_i$) of private and invariant features, then straighten them to 1-D feature vectors, thus the orthogonality loss between these vectors can be calculated, which is defined as $L_{g}$.

The process of $L_{f}$ and $L_{g}$ can be defined as:
\begin{equation}
    \begin{split}
        v^d_{i_t} = C^d_{rs}(f^d_{i_t})&, v^d_{p_t} = C^d_{rp}(f^d_{p_t}) \\
        v^n_{i_t} = C^n_{rs}(f^n_{i_t})&, v^n_{p_t} = C^n_{rp}(f^n_{p_t})
    \end{split}
\end{equation}
where $C^d_{rs}$, $C^d_{rp}$, $C^n_{rs}$ and $C^n_{rp}$ are the $1 \times 1$ convolution operation for invariant and private features of day-time and night-time images, respectively. 
\begin{equation}
    \begin{split}
        &L_{ortho} = L_f + L_g \\
        &L_f = V(v^d_{i_t}) \cdot V(v^d_{p_t}) + V(v^n_{i_t}) \cdot V(v^n_{p_t})  \\
        &L_g = V(\eta^d_{i_t}) \cdot V(\eta^d_{p_t}) + V(\eta^n_{i_t}) \cdot V(\eta^n_{p_t})
    \end{split}
\end{equation}
where $V$ is the operation which convert multi-dimensional features to 1-D features.








\subsubsection{Photometric Loss}

Following Monodepth2\cite{godard2019digging}, photometric loss are utilized in the self-supervised depth estimation process. Photometric loss $L_{pm}$ can be formulated as the same as \cite{godard2019digging}:

\vspace{-0.1in}
\begin{equation}
    \begin{split}
       & \hat{I}^d_{t} = P(d^d_{t}, pose^d_{(t-1,t)}, I^d_{(t-1)}) \\
        & \hat{I}^n_{t} = P(d^n_{t}, pose^n_{(t-1,t)}, I^n_{(t-1)}) \\
         & L_{pm} = \frac{\alpha}{2}(1-{\rm SSIM}({\hat{I}^d_{t}}, I^d_{t})) + (1-\alpha)\|{\hat{I}^d_{t}} - I^d_{t}\|_1 \\ 
         & + \frac{\alpha}{2}(1-{\rm SSIM}({\hat{I}^n_{t}}, I^n_{t})) + (1-\alpha)\|{\hat{I}^n_{t}} - I^n_{t}\|_1
    \end{split}
\end{equation}
where $pose_{(t-1,t)}$ is the pose estimation process, and following~\cite{godard2019digging}, here we use method~\cite{Direct-Methods:cvpr2018} for pose estimation, $\hat{I}$ is the reprojection image, and $\alpha = 0.85$ which is set empirically (same as \cite{godard2019digging}).



\vspace{-0.1in}
\subsubsection{Total Loss}
The total training loss of the network is 
\begin{equation}
    \begin{split}
    L_{total} = \lambda_1 L_{recons} + \lambda_2 L_{simi} +  \lambda_3 L_{ortho} + \lambda_4 L_{pm}
    \end{split}
\end{equation}
where $\lambda_1,\lambda_2,\lambda_3,\lambda_4$ are the weight parameters. In this paper, we set $\lambda_1=0.1$, $\lambda_2=\lambda_3=\lambda_4=1$, empirically.

\subsection{Inference Process}

For day-time image, the output $d^d_{t}$ is $D_s(E_s(O^d(I^d_{t})))$; for night-time $d^n_{t}$ is $D_s(E_s(O^n(I^n_{t})))$. Except for the first convolution layer, the remaining parameters of the depth estimation during the day and night are all shared. In inference, only operations of $O^d$, $O^n$ and shared weights depth network are used to estimate depth.


\section{Experiments}

\begin{table*}[htb!]
    \footnotesize
	\begin{center}
	\small
		\definecolor{gbypink}{rgb}{0.99, 0.91, 0.95}
\definecolor{gbygray}{rgb}{0.5,0.9,0.9}

\begin{tabular}{ c c c c c c c c c}
	\hline
     \rowcolor{gbypink} Method (\textbf{test at night}) &  Max depth &  Abs Rel & Sq Rel & RMSE & RMSE log & $\delta<{1.25}$ & $\delta<{1.25^2}$ & $\delta<{1.25^3}$   \\
	\hline
	Monodepth2~\cite{godard2019digging}(day) &40m & 0.477 & 5.389 & 9.163 & 0.466 & 0.351 & 0.635 & 0.826\\
    Monodepth2~\cite{godard2019digging}(night) &40m & 0.661 & 25.213 & 12.187 & 0.553 & 0.551 & 0.849 & 0.914 \\
	Monodepth2+CycleGAN~\cite{CycleGAN2017} &40m & 0.246 & 2.870 & 7.377 & 0.289 & 0.672 & 0.890 & 0.950  \\
	HR-Depth~\cite{lyu2020hr}  &40m & 0.512 & 5.800 & 8.726 & 0.484 & 0.388 & 0.666 & 0.827\\
	ADFA\footnotemark[1]~\cite{vankadari2020unsupervised} &40m &\textbf{0.201} &2.575 &7.172 &0.278 &\textbf{0.735} &0.883 &0.942 \\
	Ours    &40m &   0.233 & \textbf{2.344} & \textbf{6.859} & \textbf{0.270} & 0.631 & \textbf{0.908} & \textbf{0.962}\\
	
	\hline
	Monodepth2~\cite{godard2019digging}(day) &60m &  0.432 & 5.366 & 11.267 & 0.463 & 0.361 & 0.653 & 0.839\\
    Monodepth2~\cite{godard2019digging}(night) &60m &0.580 & 21.446 & 12.771 & 0.521 & 0.552 & 0.840 & 0.920\\
	Monodepth2+CycleGAN~\cite{CycleGAN2017} &60m &  0.244 & 3.202 & 9.427 & 0.306 & 0.644 & 0.872 & 0.946 \\
	HR-Depth~\cite{lyu2020hr} &60m &   0.462 & 5.660 & 11.009 & 0.477 & 0.374 & 0.670 & 0.842\\
	ADFA\footnotemark[1]~\cite{vankadari2020unsupervised} &60m &0.233 &3.783 &10.089 &0.319 &\textbf{0.668} &0.844 &0.924 \\
	Ours    &60m  &  \textbf{0.231} & \textbf{2.674} & \textbf{8.800} & \textbf{0.286} & 0.620 & \textbf{0.892} & \textbf{0.956}\\

	\hline
  \rowcolor{gbypink}   Method (\textbf{test at day}) &  Max depth &  Abs Rel & Sq Rel & RMSE & RMSE log & $\delta<{1.25}$ & $\delta<{1.25^2}$ & $\delta<{1.25^3}$  \\
	\hline
	Monodepth2~\cite{godard2019digging}(day) &40m & 0.117 & 0.673 & 3.747 & 0.161 & 0.867 & 0.973 & 0.991\\
	Monodepth2~\cite{godard2019digging}(night) &40m & 0.306 & 2.313 & 5.468 & 0.325 & 0.545 & 0.842 & 0.937\\
	HR-Depth~\cite{lyu2020hr}  &40m & 0.121 & 0.732 & 3.947 & 0.166 & 0.848 & 0.970 & 0.991\\
	Ours         &40m &   \textbf{0.109} & \textbf{0.584} & \textbf{3.578} & \textbf{0.153} & \textbf{0.880} & \textbf{0.976} & \textbf{0.992}\\
	
	\hline
	Monodepth2~\cite{godard2019digging}(day) &60m & 0.124 & 0.931 & 5.208 & 0.178 & 0.844 & 0.963 & \textbf{0.989}\\
    Monodepth2~\cite{godard2019digging}(night) &60m & 0.294 & 2.533 & 7.278 & 0.338 & 0.541 & 0.831 & 0.934\\
	HR-Depth~\cite{lyu2020hr} &60m &    0.129 & 1.013 & 5.468 & 0.184 & 0.825 & 0.958 & \textbf{0.989}\\
	Ours     &60m  &  \textbf{0.115} & \textbf{0.794} & \textbf{4.855} & \textbf{0.168} & \textbf{0.863} & \textbf{0.967} & \textbf{0.989}\\

	\hline

\end{tabular}
	\end{center}
	\caption{Quantitative comparison with state-of-the-art methods. Higher value is better for the last three columns, lower value is better for others. Monodepth2\cite{godard2019digging}(day) and HR-Depth\cite{lyu2020hr} mean training with day-time data of the Oxford dataset and testing on the night and day test set. Monodepth2\cite{godard2019digging}(night) means training with night-time data of the Oxford dataset and testing on the night and day test set. CycleGAN\cite{CycleGAN2017} means translating night-time Oxford images into day-time images and then using a day-time trained Monodepth2 model to estimate depth from these translated images the same as \cite{vankadari2020unsupervised}. The best results are highlighted.}
	\label{table:evaluation}
	\vspace{-0.15in}
\end{table*}


In this section, following~\cite{vankadari2020unsupervised}, we compare the performance of our method with state-of-the-art approaches on Oxford RobotCar dataset\cite{maddern20171}, which is split to adapt all day time monocular image depth estimation. 

\subsection{Oxford RobotCar Dataset}

The KITTI \cite{geiger2013vision} and Cityscapes \cite{cordts2016cityscapes} datasets are widely used in depth estimation task. However, only day-time images are contained in these datasets, which cannot meet the requirements for all-day depth estimation. Therefore, we choose the Oxford RobotCar dataset \cite{maddern20171} as the training and testing dataset. It is a large outdoor-driving dataset that contains images at various times captured in one year, including day and night times. Following~\cite{vankadari2020unsupervised}, we use the left images with the resolution of $960\times1280$ (collected by the front stereo-camera (Bumblebee XB3)) for self-supervised depth estimation. Sequence ''2014-12-09-13-21-02'' and ''2014-12-16-18-44-24'' are used for day-time and night-time training, respectively. Both training data are selected from the first 5 splits. The testing data are collected from the other splits of the Oxford RobotCar dataset, which contains 451 day-times images and 411 night-times images. We use the depth data captured by the front LMS-151 depth sensors as the ground truth in the testing phase. The images are first center-cropped to $640\times1280$, and then resized to $256\times512$ as the inputs of the network.

\subsection{Implementation Details}

First, we use CycleGAN\cite{CycleGAN2017} to translate day-time images to night-time images, then generated images and the day-time images are regarded as image pairs which are the input of the proposed domain-separated network. 




The network is trained 20 epochs in end-to-end manner with Adam optimizer ($\beta_1$=0.9, $\beta_2$=0.999). We set batch size as 6, the initial learning rate as 1e-4 for the first $15$ epochs, and the learning rate is set as 1e-5 for the remaining epochs. In inference, except for the first convolution layer, the day and night branches share weights in the depth estimation part.

\begin{figure*}[htb]
	\centering
	\includegraphics[width=1\linewidth]{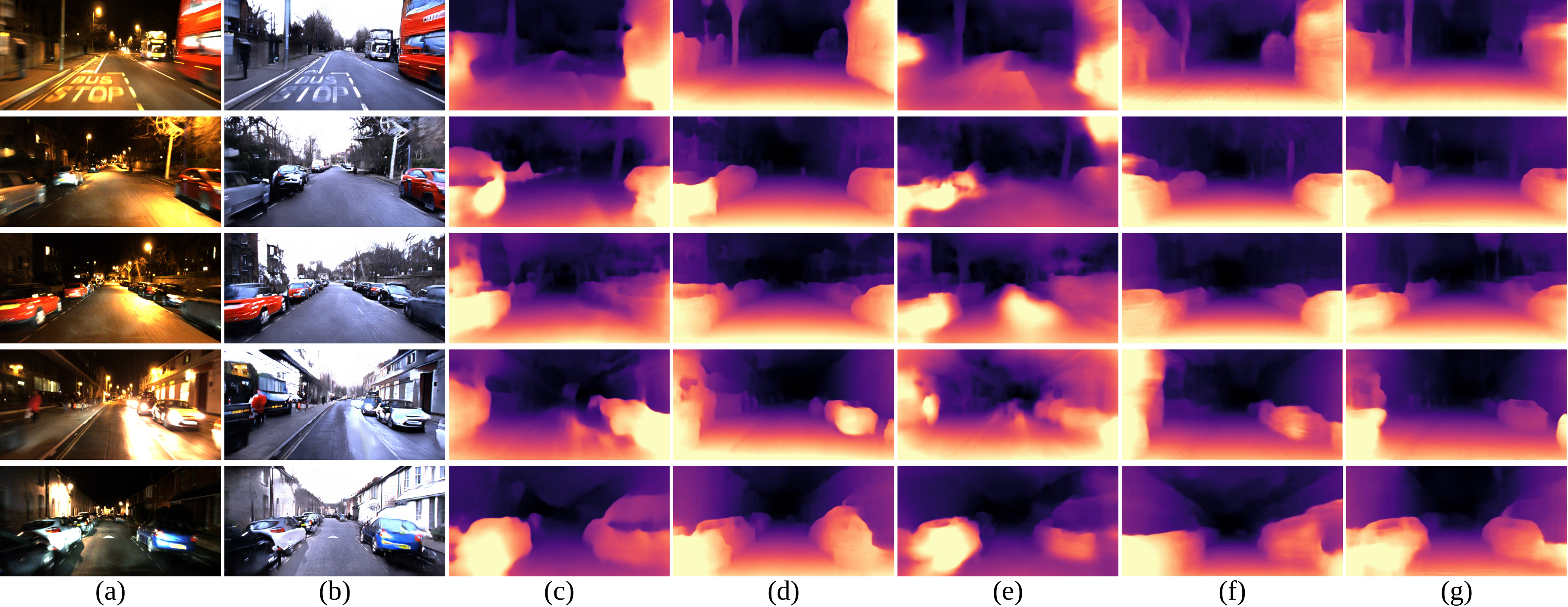}
	
	\caption{Qualitative comparison with other state-of-the-art methods at night. From left to right: (a) Night Images, (b) Fake Day Images translated by CycleGAN\cite{CycleGAN2017}, (c) Monodepth2\cite{godard2019digging}, (d) Monodepth2+CycleGAN\cite{CycleGAN2017}, (e) HR-Depth\cite{lyu2020hr}, (f) ADFA\cite{vankadari2020unsupervised}, (g) Ours. }
	\label{fig_result_night}
	\vspace{-0.2in}
\end{figure*}

\begin{figure}[htb]
	\centering
	\includegraphics[width=1\linewidth]{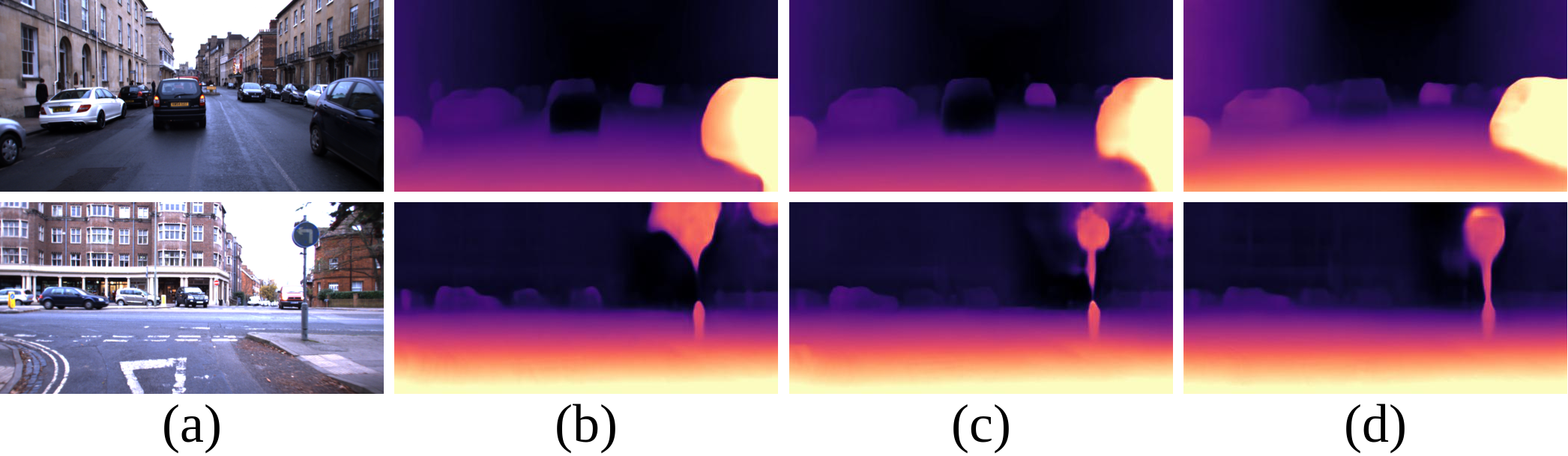}
	
	\caption{Qualitative comparison with other state-of-the-art methods at day. From left to right: (a) Day Images, (b) Monodepth2~\cite{godard2019digging}(day), (c) HR-Depth~\cite{lyu2020hr}, (d) Ours. }
	\label{fig_result_day}
	\vspace{-0.2in}
\end{figure}

\begin{figure*}[htb]
	\centering
	\includegraphics[width=1\linewidth]{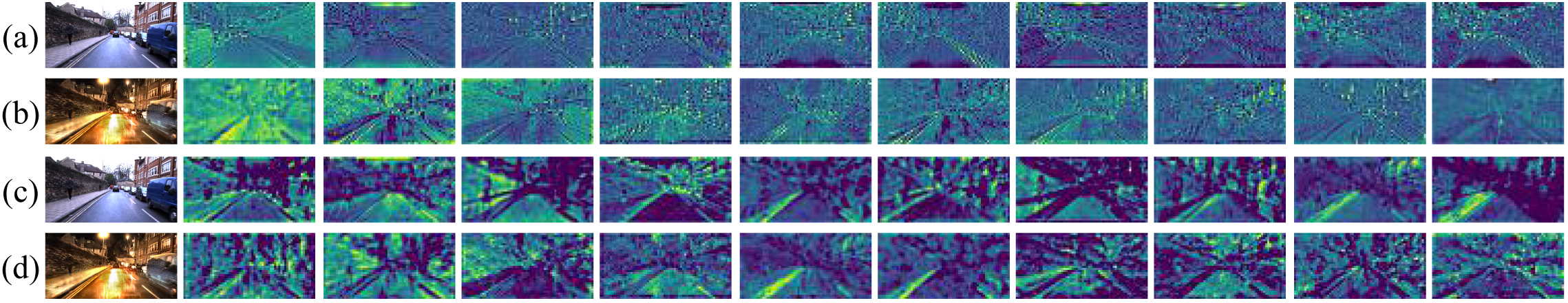}
	
	\caption{Visualization of Convolution Features. From left to right: (a) Day-time Private Features, (b)Night-time Private Features, (c)Day-time Invariant Features, (d)Night-time Invariant Features. The first column shows the corresponding input images, and the remaining columns from left to right are top $10$ feature maps that contain more information.}
	\label{fig_feature_visual}
\end{figure*}


\footnotetext[1]{Note that the test set of ADFA~\cite{vankadari2020unsupervised} is not available, our test set is not exactly the same as the test set of ADFA~\cite{vankadari2020unsupervised}.}

\subsection{Quantitative Results}


Table.~\ref{table:evaluation} demonstrates the quantitative comparison results between our approach and state-of-the-art approaches. Following~\cite{vankadari2020unsupervised}, we evaluate the performance with two depth ranges: within 40m and 60m. In Table.~\ref{table:evaluation}, Monodepth2~\cite{godard2019digging}(day) means the results trained with day-time images of the Oxford dataset, while Monodepth2~\cite{godard2019digging}(night) means the results trained with night-time images of the Oxford dataset. Monodepth2+CycleGAN~\cite{CycleGAN2017} means the results of translating night-time Oxford images into day-time images and then use a day-time trained Monodepth2~\cite{godard2019digging} model to estimate depth from these translated images the same as~\cite{vankadari2020unsupervised}.

As shown in Table.~\ref{table:evaluation}, Monodepth2~\cite{godard2019digging} is an effective self-supervised depth estimation approach, which works well for day-time images. However, the performances are limited on night-time images for all models trained by day-time and night-time images. Due to the non-uniform illumination of night-time images, areas that are too bright and too dark will cause varying degrees loss of information, which leads to the fact that training directly on images at night cannot get absolutely good results. Meanwhile, although Monodepth2+CycleGAN\cite{CycleGAN2017} and HR-Depth~\cite{lyu2020hr} can improve the depth estimation results of the night images to a certain extent, the performances are also limited due to the non-uniform illumination of night-time images. 
ADFA~\cite{vankadari2020unsupervised} reduces the domain shift between day and night images at the feature level, but the performance is limited by day-time results. As shown in Table.~\ref{table:evaluation}, our domain-separated framework can effectively relieve the influence of disturbing terms, which can improve the depth estimation performance for all-day images. Almost all performance metrics in the depth ranges of 40m and 60m for day and night images can be largely improved by our approach, which prove the superiority of our method.

\subsection{Qualitative Results}



The qualitative comparison results of night-time images are shown in Fig.~\ref{fig_result_night}, where (b) shows the translated day-times images of CycleGAN~\cite{CycleGAN2017} from night-time images, (c) shows the results of Monodepth2~\cite{godard2019digging} trained with day-time images and tested with night-time images, (d) shows the results of Monodpeth2~\cite{godard2019digging} trained with day-time images and tested with generated day-time images (CycleGAN~\cite{CycleGAN2017}). Compared with (c), it is obvious that (d) obtained better visual results, which prove that CycleGAN works positively for night-time depth estimation. (f) shows the result of ADFA~\cite{vankadari2020unsupervised} which uses generative adversarial strategy in feature level of night-time images, which cannot fully restore the contour of the objects. Comparing with (c) to (f), more visual apparelling results can be obtained by our approach, which proves the effectiveness of our method.


Fig.\ref{fig_result_day} demonstrates the qualitative comparison results of day-time images. We can see that more depth details can be recovered by our approach, which clarifies the ability of our method for day-time images. 

\begin{table*}[htb!]
     \footnotesize
	\begin{center}
		\definecolor{gbypink}{rgb}{0.99, 0.91, 0.95}
\definecolor{gbygray}{rgb}{0.5,0.9,0.9}

\begin{tabular}{ ccccc c c c c c c c c}
	
	\hline
     \rowcolor{gbypink} Method (\textbf{night}) & Paired &  $L_{recons}$ &  $L_{f}$ & $L_g$ & $L_{simi}$ &  Abs Rel & Sq Rel & RMSE & RMSE log & $\delta<{1.25}$ & $\delta<{1.25^2}$ & $\delta<{1.25^3}$   \\
	\hline
	
      $U$    &    & & & & &        0.573 & 18.577 & 11.189 & 0.524 & 0.569 & 0.807 & 0.897 \\
	   $P$   &$\surd$ & & & &      &  0.429 & 15.183 & 11.401 & 0.422 & 0.589 & 0.862 & 0.942 \\
	
	$PR$ &$\surd$ &$\surd$ & & & &  0.357 & 10.699 & 10.385 & 0.377 & 0.611 & 0.884 & 0.946\\
	$PRF$ &$\surd$ &$\surd$ &$\surd$ & & &  0.251 & 2.993 & 8.173 & 0.299 & 0.606 & 0.884 & 0.949\\
	$PRFG$  &  $\surd$  &$\surd$ & $\surd$&$\surd$ & &   0.231 & 2.453 & 7.327 & 0.282 & 0.662 & 0.900 & 0.956 \\

	$PRFGS$     &$\surd$ &$\surd$ &$\surd$ &$\surd$ &  $\surd$  & 0.233 & 2.344 & 6.859 & 0.270 & 0.631 & 0.908 & 0.962  \\
    
    \hline
   \rowcolor{gbypink} Method (\textbf{day}) & Paired &  $L_{recons}$ &  $L_{f}$ & $L_g$ & $L_{simi}$ &  Abs Rel & Sq Rel & RMSE & RMSE log & $\delta<{1.25}$ & $\delta<{1.25^2}$ & $\delta<{1.25^3}$   \\
	\hline
	
    $U$           & & & & &  &  0.280 & 4.509 & 6.359 & 0.297 & 0.661 & 0.873 & 0.949 \\
	$P$  &  $\surd$ &  &  &  &  & 0.117 & 0.758 & 3.737 & 0.170 & 0.874 & 0.967 & 0.988 \\
	
	$PR$ &$\surd$ &$\surd$ & & & & 0.131 & 1.355 & 3.937 & 0.178 & 0.848 & 0.967 & 0.990 \\
	$PRF$ &$\surd$ &$\surd$ &$\surd$ & & &  0.108 & 0.569 & 3.535 & 0.152 & 0.879 & 0.977 & 0.992\\
	$PRFG$ & $\surd$& $\surd$&$\surd$ &$\surd$ &  &  0.109 & 0.580 & 3.518 & 0.152 & 0.891 & 0.976 & 0.991 \\
	
     $PRFGS$  &$\surd$ &$\surd$ &$\surd$ &$\surd$ & $\surd$  &   0.109 & 0.584 & 3.578 & 0.153 & 0.880 & 0.976 & 0.992 \\

	\hline

\end{tabular}

	\end{center}
	\caption{The table shows the quantitative results of the proposed losses and input data. All experiments are tested within the depth range of 40m.}
	\label{table:ablation_study}
	\vspace{-0.2in}
\end{table*}

\subsection{Ablation Study}

\subsubsection{Private and Invariant Features}

Fig.~\ref{fig_feature_visual} demonstrates the private and invariant features obtained by the private and invariant feature extractors of day and night-time images, where the first column shows the corresponding input images, and the remaining columns from left to right are top $10$ feature maps that contain more information. Row (a) and (b) are private features of the day and night-time images, while row (c) and (d) are corresponding invariant features. It is obvious to see that feature maps in Fig.~\ref{fig_feature_visual} (a) and (b) contain non-regular and smooth information with fewer structures, which is similar to the illumination information of images. Feature maps in Fig.~\ref{fig_feature_visual} (c) and (d) contain regular and texture information with obvious structures, which can represent the invariant of scenes, proving that our approach can separate the private (illumination, etc.) and invariant (texture, etc.) information effectively.

\subsubsection{Analysis of Input Data and Losses}

\textbf{Unpaired data vs. paired data:} The results of $U$ and $P$ in Table.~\ref{table:ablation_study} show the quantitative results of our method with unpaired and paired images as input. In specific, we use the day-time and night-time images captured in the same roads from Oxford RobotCar dataset as unpaired data ($U$), and we use day-time images and the corresponding generated night-time images (generated by GAN) as paired data ($P$). We can see that results obtained by paired data outperform results obtained by unpaired data, which is mainly because that inconsistent information exists in unpaired images since they are captured at different times, though on the same roads. Hence, we use paired images in this paper.



\textbf{Reconstruction loss:} The private and invariant features are complementary, which should contain all the information of the original images. Therefore, we use reconstruction loss. Table.\ref{table:ablation_study} $PR$ shows the quantitative results of our approach with reconstruction loss. Comparing with $P$, $PR$ produces great improvement to the depth estimation of night-time images. However, the result of the day-time images is slightly worse, because the invariant feature extractor is shared during the day and night, and no other constraints are used.




\textbf{Orthogonality loss:} Orthogonality loss is used to guarantee that private and invariant features can be separated by the private and invariant extractors orthogonally, which is composed of $L_f$ and $L_g$. Table.\ref{table:ablation_study}. $PRF$ and $PRFG$ show the quantitative results of our approach with $L_f$ and $L_g$, respectively. Compared with $PR$, $PRF$ (added $L_f$ loss) can greatly improve the performance of depth estimation night-time images, which also works positively for day-time images. Meanwhile, $PRFG$ (added $L_g$ loss) can help the network achieves better performance on night-time images while maintaining the performance of day-time images, thereby further improving the performance of depth estimation for all-day images. 


\textbf{Similarity loss:} The estimated depth maps should be similar for the input paired images, because consistent information is contained. Hence, similarity loss is employed in our approach. Since the depth estimation process of the day-time usually achieves better results than night, we use the day-time depth as a pseudo label so that the depth of the paired night image should be close to the day-time. Table.\ref{table:ablation_study}. $PRFGS$ demonstrates the quantitative results of our approach with similarity loss, which shows that the similarity constraint can further improve the depth estimation result of nigh-time images while maintaining the performance of day-time images, thus proving the effectiveness of the similarity loss.






\section{Conclusion}

In this paper, to relieve the problem of low-visibility and non-uniform illumination in self-supervised depth estimation of all-day images, we propose an effective domain separated framework, which separates the images into two complementary sub-spaces in feature levels, including private (illumination, etc.) and invariant (texture, etc.) domains. The invariant (texture, etc.) features are employed for depth estimation, thus the influence of disturbing terms, such as low-visibility and non-uniform illumination in images, can be relieved, and effective depth information can be obtained. To alleviate the inconsistent information of day and night images, the domain-separated network takes the day-time images and the corresponding generated night-time images (GAN) as input. Meanwhile, orthogonality, similarity and reconstruction losses are utilized to separate and constrain the private and invariant features effectively, thus better depth estimation results can be expected. Note that the proposed approach is fully self-supervised and can be trained end-to-end, which can adapt to both the day and night images. Experiments on the challenging Oxford RobotCar dataset demonstrate that our framework achieves state-of-the-art results for all-day images.  



\noindent\textbf{Acknowledgements} This work is supported in part by Robotics and Autonomous Driving Lab of Baidu Research. This work is also supported in part by the National Key R\&D Program of China under Grant (No.2018YFB1305900) and the National Natural Science Foundation of China under Grant (No.61836015).


{\small
\bibliographystyle{ieee_fullname}
\bibliography{egbib}
}

\end{document}